\pdfoutput=1

\documentclass[11pt]{article}

\usepackage[preprint]{acl}

\usepackage{multirow}
\usepackage{times}
\usepackage{latexsym}
\usepackage{amsfonts,amssymb}
\usepackage{amsmath}
\usepackage{mathtools}
    \mathtoolsset{showonlyrefs, showmanualtags, mathic}
\usepackage{mleftright}
\usepackage{subcaption}
\usepackage{xurl}

\usepackage{zref-clever}
	\zcsetup{
		cap=true,
		abbrev=false
	}
\usepackage{zref-titleref}

\usepackage[T1]{fontenc}

\usepackage[utf8]{inputenc}

\usepackage{microtype}

\usepackage{inconsolata}

\usepackage{graphicx}

\usepackage{booktabs}

\usepackage{csquotes}
\usepackage{cleveref}
\usepackage{siunitx}
	\DeclareSIUnit{\quantity}{\relax}
	\DeclareSIUnit{\words}{words}
	\DeclareSIUnit{\sentences}{sentences}
\usepackage{tabularray}
    \UseTblrLibrary{booktabs}
    \UseTblrLibrary{siunitx}

\DeclarePairedDelimiter\abs{\lvert}{\rvert}

\title{On the scaling relationship between cloze probabilities and language model next-token prediction}

\author{
    Cassandra L. Jacobs \\
    Department of Linguistics \\ %
    University at Buffalo \\
    Buffalo, NY, USA \\
    \texttt{cxjacobs@buffalo.edu} \\\And
    Morgan Grobol \\
    MoDyCo\\
    Université Paris Nanterre \\
    Paris, France\\
    \texttt{lgrobol@parisnanterre.fr}
}

\date{}

\begin{document}
\maketitle
\begin{abstract}
Recent work has shown that larger language models have better predictive power for eye movement and reading time data.
While even the best models under-allocate probability mass to human responses, larger models assign higher-quality estimates of next tokens and their likelihood of production in cloze data because they are less sensitive to lexical co-occurrence statistics while being better aligned semantically to human cloze responses.
The results provide support for the claim that the greater memorization capacity of larger models helps them guess more semantically appropriate words, but makes them less sensitive to low-level information that is relevant for word recognition.
\end{abstract}

\section{Introduction}

Humans regularly engage in linguistic prediction, and can often guess what others are about to say.
The predictability of a word in context is often quantified by computing its \textit{cloze probability}: the proportion of human participants who provide it when completing a sentence in the cloze task (e.g., \enquote{He hated bees and feared encountering a \_\_\_}; \citealp{taylor1953cloze}).
Cloze probabilities partly reflect a word's appropriateness in a given linguistic context or situation, and are determined in part by semantic fit to the sentence and knowledge of word and multiword statistics \cite{smith2011cloze}.
In human experiments, words with higher cloze probabilities are typically read more quickly and/or trigger more attenuated neural activity \cite{de2024cloze}.
However, there is ongoing disagreement about the statistical and psycholinguistic predictive properties of cloze data, which are sparse and challenging to collect, and recent works have turned to probabilities computed using large language models as a replacement proxy for appropriateness \cite{shain2024large,nair2026clozinggapexploringlanguage}.

While we know that large language models diverge meaningfully from human production preferences \cite{eisape-etal-2020-cloze,kendro2026large,nair2026clozinggapexploringlanguage,pivelvillanueva2026llmscapturestablehumangenerated}, it is unclear what model architectures, capacities, or memorization abilities promote greater or lesser alignment to cloze data.
Larger models are more sensitive to semantics and less sensitive to n-gram statistics than smaller ones \cite{michaelovlanguage}.
Earlier work analyzing the alignment between statistical (\num{5}-gram) estimates and human cloze productions showed that humans are biased toward guessing words that are higher-frequency and more semantically similar to other words in the preceding context \citet{smith2011cloze}.
It seems reasonable to expect, then, that larger models will demonstrate improved capacity to model language production biases \cite{pickering2013integrated,smith2011cloze}.

Understanding how language models align with cloze data is especially important given the high cost of data collection for probability estimation with human subjects \cite{pivelvillanueva2026llmscapturestablehumangenerated}.
Moreover, recent analyses of reading time data have typically shown that smaller neural language models (NLMs) produce better estimates of a word's processing difficulty than cloze probabilities \cite{baroni2014don,goodkind2018predictive,shain2024large}.
Critically, language model probabilities are especially useful for words that were never guessed as continuations of the sentence \cite{nair2026clozinggapexploringlanguage,szewczyk2022context}.
Accurately predicting words that are produced by humans in cloze tasks is still useful for accounting for words that \textit{are} guessed as potential completions \cite{de2024cloze,szewczyk2022context}.

In this work, we first characterize the scaling relationship between models and the lexical alignment between model generations and cloze productions.
The results stand in contrast to similar investigations of reading time data, showing that larger models provide a \textit{better} match to cloze data.
However, better fit does not necessarily mean a \emph{good} fit.
Thus, a second goal of the present paper is to describe the degree of semantic alignment between NLM generations and human predictions.

\begin{table}
    \centering
    \caption{Top \num{5} Pythia next-subword continuations in different Pythia models for the sentence fragment, \enquote{He hated bees and feared encountering a \_\_\_\_}. $*$ \enquote{was} possibly corresponds to the first subword of \enquote{wasp} and \enquote{h} to \enquote{hive}.}
    \label{table:hive}
    \begin{tabular}{ccc|c}
        \toprule
        Pythia-70M & -160M & -2.8B & Human \\
        \midrule
        lot & swarm & swarm & hive \\
        swarm & bee & bee & swarm \\
        bee & new & h* & bee \\
        new & pest & was* & nest \\
        threat & h* & colony & wasp \\
        \bottomrule
    \end{tabular}
\end{table}

Our experiments compare cloze probabilities reported in a large-scale dataset \cite{peelle2020completion} against probabilities from NLMs trained to perform next-token prediction (Table \ref{table:hive}).
Inspired by similar work assessing the role of model capacity on fit to reading time data by \citet{oh-schuler-2023-transformer}, we leverage the Pythia suite of language models \citep{biderman2023pythia} to understand whether and when language models capture human production variability in the cloze task.
Pythia models are next token predictors trained on the same dataset (the Pile; \citealt{gao2020pile}) and provide a way to test the influence of model hyperparameters on fit to cloze probabilities.

In \ztitleref{sec:exp-rank}, we demonstrate the overall strengths and weaknesses of the best fitting model to human cloze probabilities (Pythia-2.8B-deduped).
\ztitleref{sec:exp-capacity} evaluates the contribution of memorization to fit to human cloze data.
We compare n-gram statistics to both NLM predictions and cloze probabilities.
Further analyses show that model size, training budget, and data deduplication all influence fit to cloze probabilities.
Finally, \ztitleref{sec:exp-clusters} probes the correspondence in semantic similarity structure between human and NLM predictions.

\section{Cloze data}

We conduct our experiments on the \citet{peelle2020completion} completion norms dataset. 
It consists of \num{3085} English sentence stems created by experimental psycholinguists for which human participants were asked to type the next word, with each stem receiving at least \num{100} manually validated responses to produce reliable cloze probability estimates.
The stimuli were constructed to vary in the degree of final-word predictability and are unlikely to be included in the Pile, in contrast to naturalistic materials \cite{luke2018provo,oh2025inverse}.
Of the \num{3085} sentences, only \num{16} sentences were shorter than \num{7} or longer than \num{10} words.
Despite the short contexts, NLMs typically guessed appropriate words, and statistical tests confirmed that sentence length did not have a significant influence on Spearman rank correlation to human cloze probabilities (\(p = .27\)).

For analyses of language model probabilities, we choose to use the first subword of human cloze responses.
While a human response might consist of several subword tokens, considering only word-initial subwords provides enough precision for our purpose while greatly simplifying probability estimation (e.g., \citealp{holtzman2021surface, giulianelli-etal-2023-comes,oh-schuler-2024-leading,nair-resnik-2023-words,pimentel-meister-2024-compute,pivelvillanueva2026llmscapturestablehumangenerated}).
This decision is also justified by the distribution of human responses: approximately \qty{50.4}{\percent} of human response tokens are very common words that correspond to a single token in the models' vocabularies.
On average, each unique response type was broken into ($1.64 \pm 0.67$) subwords.
For analyses of semantic spaces, we mean-pool subword embeddings of multitoken continuations \cite{giulianelli2024generalized,nair2026clozinggapexploringlanguage}.

\section[Experiment 1]{Experiment 1: Comparing production probabilities}\label{sec:exp-rank}

This analysis characterizes the best-case scenario for probabilistic alignment among the Pythia models by quantifying the correlation between language model-allocated probability mass and human cloze probabilities.
We show that the performance of NLMs in matching human production preferences can be impressive, but illustrate some deficiencies in the allocation of probability mass across several transformations of language model probabilities.

Pythia-2.8B-deduped obtained the overall highest correlation to the human responses.
The subsequent section will explore how model parameterization, training time, and data deduplication influence alignment.

\begin{figure*}
    \centering
    \begin{subfigure}[t]{0.475\textwidth}
        \centering
        \includegraphics[width=\textwidth]{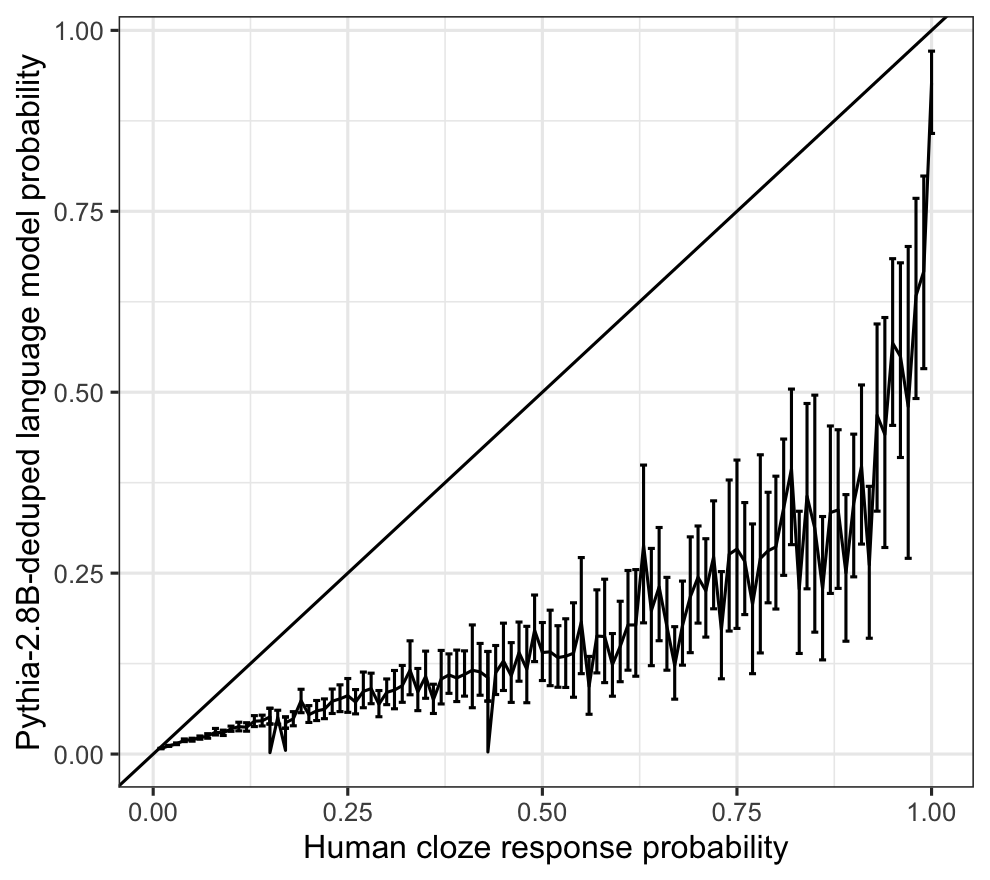}
        \caption{Relationship between language model probabilities and cloze probabilities. Error bars represent bootstrapped confidence intervals. Solid line represents 1:1 relationship.}
        \label{fig:prob-corr-plot}
    \end{subfigure}
    \hfill
    \begin{subfigure}[t]{0.475\textwidth}
        \centering
        \includegraphics[width=\linewidth]{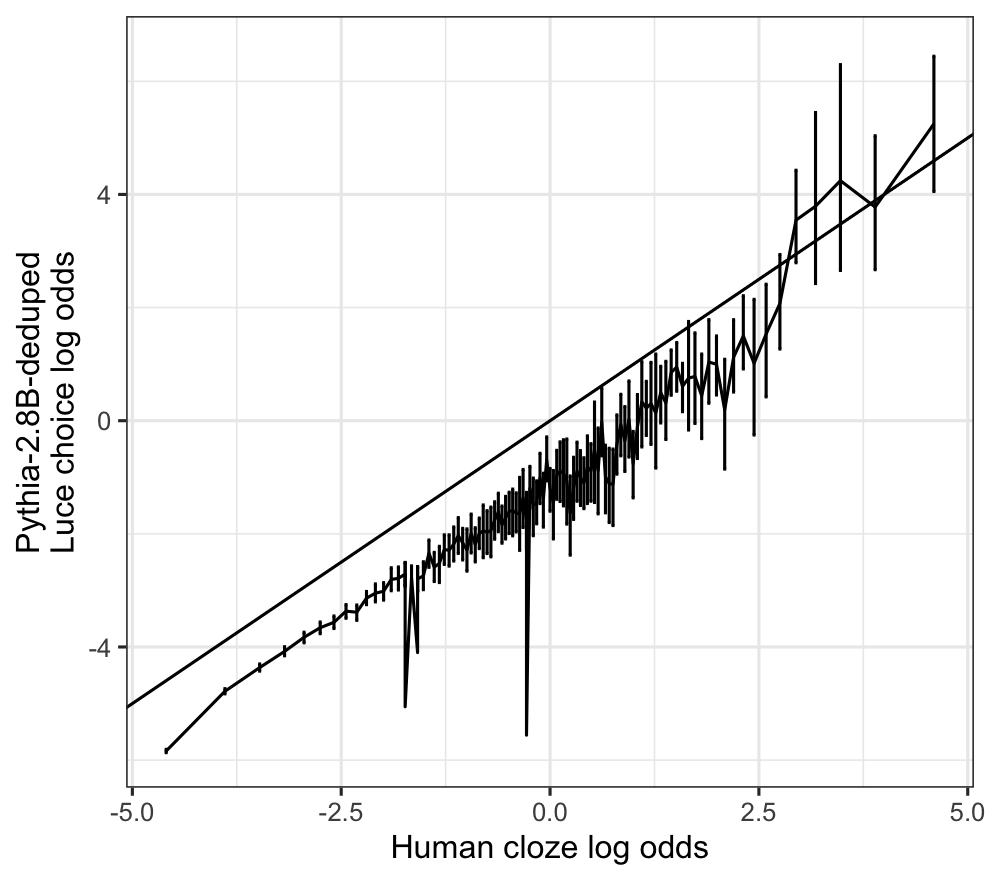}
        \caption{Relationship between log-odds of human cloze probabilities and NLM probabilities renormalized over human responses.}
        \label{fig:luce-logodds}
    \end{subfigure}
    \label{fig:prob-rank-corr-plot}
    \caption{Human-NLM cloze correlations for Pythia-2.8B-deduped showing major deficiencies in assigning probability mass to human completions. All models at all sizes showed similar patterns and are not pictured here.}
\end{figure*}

\subsection{Cloze probabilities}\label{sec:cloze-probs}

Pythia probabilities and Peelle cloze probabilities are highly correlated with each other (Pearson's \(\rho = 0.492\); Spearman's \(\rho = 0.485\)), but this correlation is complex (\zcref{fig:prob-corr-plot}).
Firstly, many cloze completions are assigned probabilities approximately half of their empirical probability in the corpus.
For instance, glancing at Figure \ref{fig:prob-corr-plot}, it is clear that words at \num{.75} cloze probability -- representing on average \num{3} of every \num{4} responses -- were assigned around language model probabilities of \num{.25}.
Nevertheless, on average, words at the endpoints of the cloze probability scale ($p=.01$ and $p=1$, respectively) are reasonably well captured; the middle range of probabilities is, however, systematically under-estimated by language model probabilities. 
This seems to be a case of miscalibration, a common issue in neural network classifiers \citep{Guo2017OnCO}. 

\subsection{Cloze log-odds}
Comparing raw probabilities, as in \zcref{sec:cloze-probs}, is made inconvenient by the skewness of the probability distributions observed.
To alleviate this, we complement them by analyses of log-odds, i.e.,\ after applying a logit transform to all probabilities \(p\) with a small smoothing parameter \(\alpha\):
\begin{equation}
    \mathrm{logit}(p) = \log\frac{p + \alpha}{1-(p + \alpha)}, \alpha=\num{1e-06}
\end{equation}

Unlike raw probabilities, log-odds are unbounded, which helps to make low-probability tails less skewed. 
The results of a linear mixed effects regression with random intercepts by sentence stem and cloze response reveal a correlation between language model log-odds and cloze probabilities that was significantly positive ($\beta = 0.244$, $t(42410)=133.9$, $p < .001$).
The large negative estimate of the intercept term reveals that language model probability mass is assigned more uniformly ($\beta = -2.275$, $t(23930)=-167.6$, $p < .001$).

For a cleaner comparison to human responses, which sum to \num{1}, we also renormalize the NLM probabilities for each preamble using Luce's choice rule \cite{luce1977choice}, instead of taking the raw value after softmax over the NLM's whole vocabulary.
The equivalent analysis with Luce choice log-odds analysis revealed that while it is possible to produce a stronger correlation to cloze probabilities ($\beta = 0.252$, $t(50440)=139.7$, $p < .001$), Pythia still mostly under-allocates probability mass as seen in the intercept term ($\beta = -2.572$, $t(18290)=-226.7$, $p < .001$; left side of \zcref{fig:luce-logodds}).
At the very highest levels of cloze probability (on the right side of the figure), Luce-choice language model log-odds may slightly over-allocate probability mass.
These results provide converging evidence that accounting for differences in the response space does not eliminate the tendency of Pythia-2.8B-deduped to under-allocate probabilities.

\subsection{Rank correlations}

\begin{figure}[th]
    \centering
    \includegraphics[width=.9\linewidth]{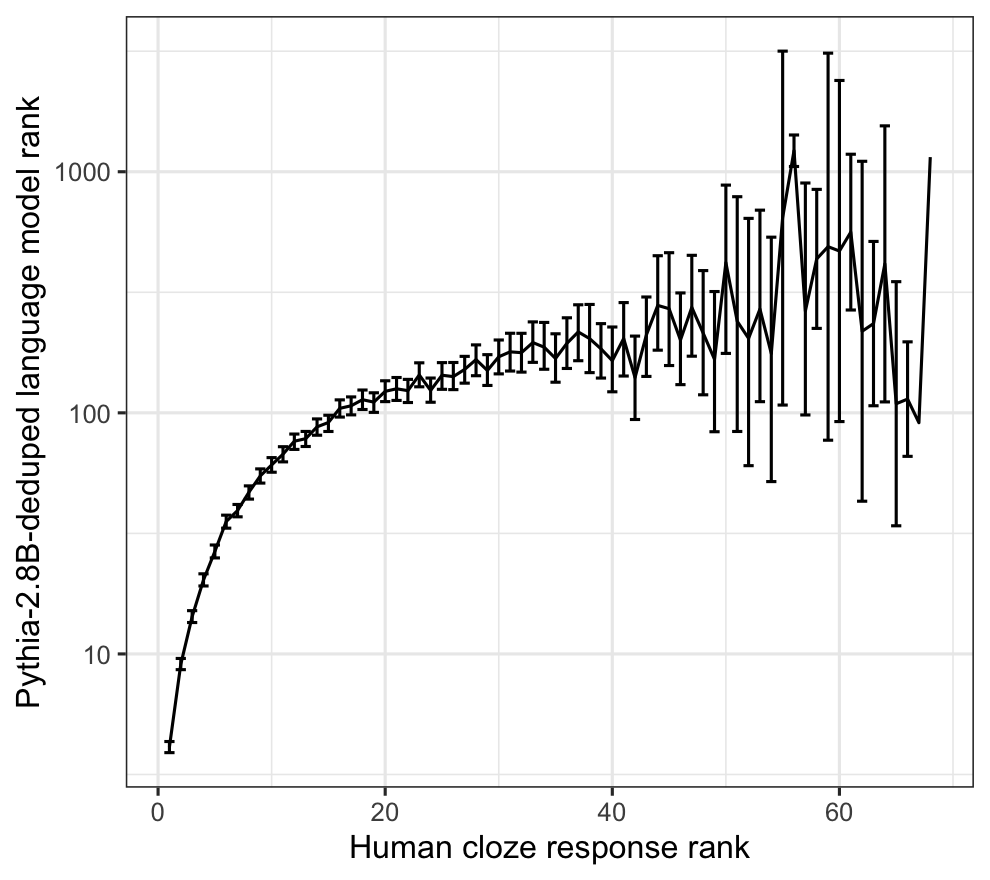}
     \caption{Relationship between language model ranks and cloze ranks. Error bars represent bootstrapped confidence intervals.}
     \label{fig:rank-corr-plot}
\end{figure}

Both human cloze and NLM next-token probabilities are typically skewed, with a long tail of rare responses.
In this analysis, we follow \citet{sinha2023language} and compare the NLM \emph{rank} of a human response's first subword probability against the empirical rank of the response in the cloze dataset.
This allows us to constrain the correlation to the same numerical range and determine if the language models are able to accurately order human responses, as in a multiple-choice test \cite{holtzman2019curious}.
This transformation partly addresses the skew in human probabilities and ensures an even distribution of values even when the vocabulary sizes are quite different.

We measure the alignment between human and LM rank distributions with \citeposs{spearman1904ProofMeasurementAssociation} rank correlation coefficient \(\rho\).
Consistent with the previous analyses, the human cloze and language model probabilities are strongly correlated ($\rho = 0.48$, $p < .001$).
However, one need not be optimistic about ranks: while there is a clear correspondence between NLM and human cloze probabilities, the model consistently under-ranks probable human responses (\zcref{fig:rank-corr-plot}).

\section[Experiment 2]{Experiment 2: Model capacity and human cloze alignment}\label{sec:exp-capacity}

The previous experiment established that models systematically under-allocate the probability mass of human responses in the cloze task.
This experiment builds on findings that language model next-token predictions are strongly correlated with unigram probabilities, n-gram probabilities, and (contextual) semantic similarity \cite{michaelovlanguage}.
Thus, we further examine how the correlations between NLM outputs from Pythia models of all sizes (14M, 70M, 160M, 410M, 1B, 1.4B, and 2.8B), trained on the Pile and the de-duplicated Pile compare to cloze data and simple n-gram statistics.

\subsection[Experiment 2.1]{Pythia correlation to n-gram baseline}\label{sec:exp-ngram-capacity}
This experiment probes potential sources of the correlation between language model probabilities and human cloze responses in Section \ref{sec:exp-rank}, while laying the groundwork for our comparison of NLMs of varying sizes and training durations in Section \ref{sec:exp-pythia}.
Here we examine the correlation between the ranks of n-gram-based next-token prediction scores to the ranks of language model and human cloze probabilities in the \citet{peelle2020completion} norms.

Following \citet{michaelovlanguage}, we estimated scores for a 5-gram model smoothed with \citeposs{brants2007LargeLanguageModels} Stupid Backoff heuristic (\(n=5\), \(\alpha=0.4\)) on Wikipedia\footnote{Precisely, the 2023-11-01 dump of the English version of Wikipedia, tokenized with the GPT-2 tokenizer.} and found strong correlations with NLM next-token probabilities.
Interestingly, model capacity and data deduplication modulated the correlation between language model probabilities and 5-gram scores.
We find that the Pythia-70M model produced the strongest correlation to 5-gram estimates ($\rho = 0.57$), and this relationship decreased in magnitude as models get larger; Pythia-2.8B model showed a much weaker correlation to 5-gram statistics ($\rho = 0.44$).

\begin{figure}
    \centering
    \includegraphics[width=.85\linewidth]{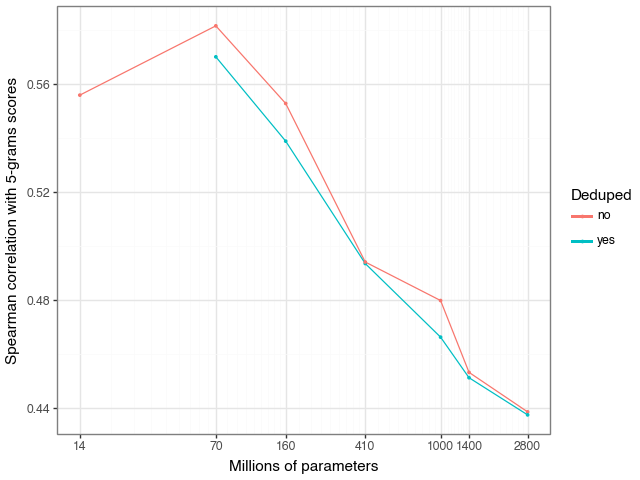}
    \caption{Correlation strength between Wikipedia 5-gram scores and Pythia next-token probabilities by model size and data deduplication. X axis in log scale.}
    \label{fig:ngram-llm-spearman-parameters}
\end{figure}

Data deduplication in the Pile negatively impacts the correlation between Wikipedia 5-gram scores and language model probabilities.
Additionally, consistent with prior findings showing that NLMs rely less on n-gram statistics as they become larger, we find that the effect of deduplication on correlation to 5-gram scores becomes smaller as models get larger (\zcref{fig:ngram-llm-spearman-parameters}).
These results suggest that smaller NLMs are more prone to memorization of n-gram statistics, and data deduplication reduces the bias toward these repeated sequences.

NLM probabilities were only weakly correlated with unigram scores, which were themselves strongly correlated with 5-gram scores.
Human cloze responses showed a weak correlation with 5-gram scores and very weak correlation to unigram scores.
We summarize these correlations in Table \ref{tab:ngram-lm-human-correlations}.
Of particular note is the size of the correlation between NLM probabilities and 5-gram scores, which is similar in magnitude to the correlation between the best NLM and human probabilities. 

\begin{table}
    \centering
    \caption{Spearman correlation coefficient estimates between different language models (5-gram vs. Pythia NLM) and human cloze completions. All estimates significant. }
    \label{tab:ngram-lm-human-correlations}
    \begin{tblr}{
        colspec={
            c
            S[table-format=1.3]
            S[table-format=1.3]
            S[table-format=1.3]
        },
        row{1}={guard, font=\bfseries},
        column{1}={guard, font=\bfseries},
    }
        \toprule
                & Human & NLM  & 5-gram\\
        NLM     & .375  &      & \\
        5-gram  & .177  & .483 & \\
        Unigram & .007  & .187 & .445\\
        \bottomrule
    \end{tblr}
\end{table}

\subsection[Experiment 2.2]{Pythia correlation to cloze completions}\label{sec:exp-pythia}

\begin{figure}[t]
     \centering
     \begin{subfigure}[b]{.425\textwidth}
         \centering
        \includegraphics[width=\linewidth]{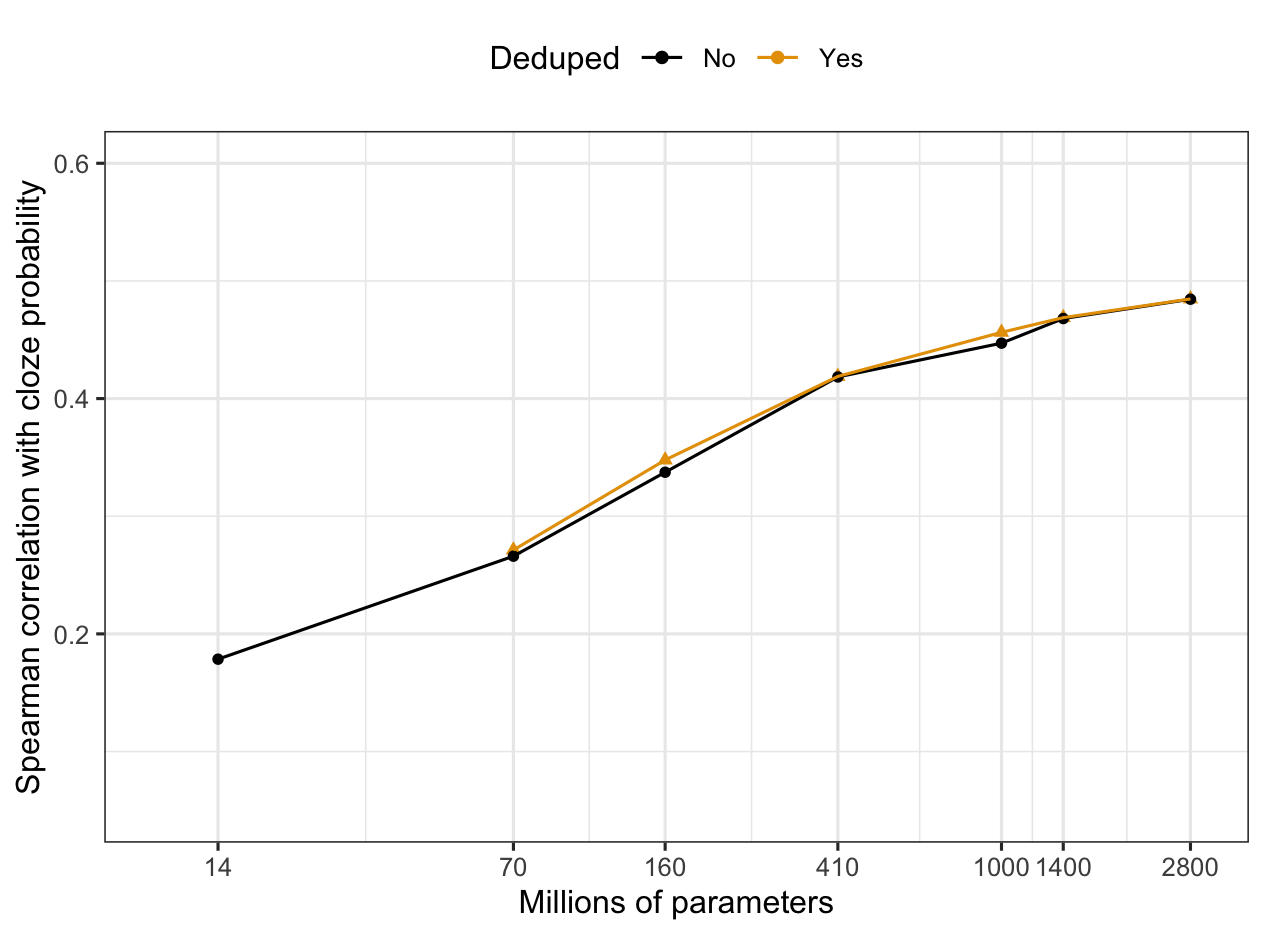}
        \caption{Effect of model size in millions of parameters on rank correlations with human responses. X axis in log scale. Note: 14M model does not have a deduplicated implementation.}
        \label{fig:pythia-nparams-spearman1}
     \end{subfigure}
     \hfill
     \begin{subfigure}[b]{.45\textwidth}
         \centering
        \includegraphics[width=.8\linewidth]{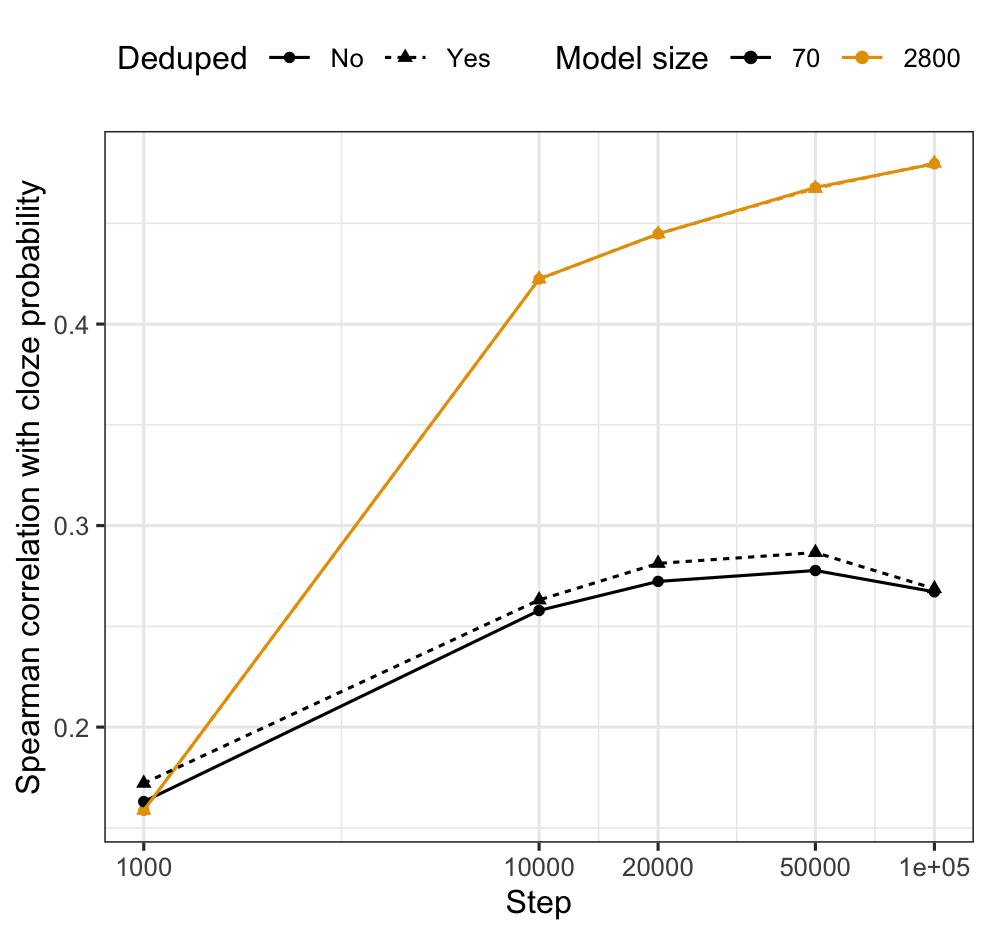}
        \caption{Effect of training budget (number of steps) on rank correlations with human responses. Deduplication plays a larger role in fit with smaller models. X axis in log scale.}
        \label{fig:pythia-nsteps-spearman}
     \end{subfigure}
    \label{fig:pythia-budget-corr}
    \caption{Effect of model capacity and size on cloze correlation.}
\end{figure}

This experiment assesses the contribution of training in NLMs to fit to human data. 
The analyses in the previous section assessed the fit between NLMs and cloze data overall, but did not closely examine the effect of model size, training duration, or data deduplication.
Better understanding the contribution of these factors is important because larger models or models trained for longer appear to produce less predictive next-word probabilities than smaller NLMs \cite{oh-schuler-2023-transformer,oh-schuler-2023-surprisal,shain2024large} but this relationship is not necessarily expected for cloze data.
We again standardize the correlation between human cloze and NLM probabilities and only consider the relative ranks of each. %
For comparisons based on model size, we use all the models up to and including Pythia-2.8B, in standard and deduplicated versions.
For comparisons based on training progress, we use checkpoints of Pythia-70M and Pythia-2.8B models with and without deduplication. 

Here performance generally improved with model size (Figure \ref{fig:pythia-nparams-spearman1}), such that greater model sizes had higher correlations with human responses.
As evidenced by the non-linear slope plotted in Figure \ref{fig:pythia-nsteps-spearman}, we observe diminishing returns over the course of multiple epochs, suggesting that longer pre-training does not necessarily lead to better estimation of human predictions.
The declining fit evident at the longest training times replicates some prior work suggesting a phase change in model behavior indicating overtraining \cite{biderman2023pythia}.
Models trained on deduplicated text typically produced higher-quality estimates of cloze rankings, even for models of the largest size (\(\Delta\mathrm{LL}=19\); \(\chi^2(0)=37.154\); \(p < 0.001\)). %

\section*{Interim summary}
\ztitleref{sec:exp-rank} showed modest-to-high correlations between human- and machine-generated responses in a cloze task for approximating rankings of human responses.
Those analyses replicated previous findings demonstrating deficiencies in next-word prediction even among the largest pretrained neural language models \cite{holtzman2019curious,vaidya2023humans,botch2024humans,klein2024effect,oh-etal-2024-frequency}.
\ztitleref{sec:exp-capacity} demonstrated a novel result that data deduplication has an important effect on the degree of correspondence between human cloze and NLM predictions, such that deduplication appears to improve the probability estimates of some combinations.
This finding coupled with the reverse relationship between Wikipedia 5-gram scores and deduplicated NLM probabilities provides additional support for the claim that memorization of n-grams is pervasive in smaller language models \cite{michaelovlanguage}.

\section[Experiment 3]{Experiment 3: Human-LM semantic representational similarity analysis}\label{sec:exp-clusters}

\begin{table*}[t]
    \centering
    \caption{Top five most similar words to the word \emph{bee} in human and NLM cloze completion semantic spaces.}\label{tab:co-occurrences}
    \begin{tblr}{
        colspec={X[c] X[c] X[c] X[c] X[c]}
    }
        \toprule
        \SetCell[c=2]{c} {Count-based PPMI + PCA} & & \SetCell[c=3]{c} {Averaging Contextual Embeddings}\\
        \cmidrule[r]{1-2}\cmidrule[l]{3-5}
        Pythia-160M & Human & Pythia-160M & Pythia-2.8B & Human\\
        \cmidrule[r]{1}\cmidrule[r,l]{2}\cmidrule[r,l]{3}\cmidrule[r,l]{4}\cmidrule[l]{5}
        mosquito  & fly  & moth &   moth & moth \\
        sting  & ant   &  insect & mosquito & mosquito \\
        snake &  mosquito & bug & insect & spider \\
        spider & insect&  spider & bug &  bug \\
        bug  & butterfly & mouse & spider &  bird \\
        \bottomrule
    \end{tblr}
\end{table*}

\citet{michaelovlanguage} argue that larger models can encode \enquote{more complex relationships} between words in text, such as anticipating (distributionally) semantically similar words.
We therefore expect that greater model capacity will promote higher semantic alignment to human cloze responses, which should be reflected in the structure of their semantic spaces.
We wish to specifically quantify comparisons such as those presented in \zcref{tab:co-occurrences}.
We explore this by comparing the distributional semantic spaces from human and language model-generated sources using a procedure known as Representational Similarity Analysis (RSA; \citealp{kriegeskorte2008representational}).
For all model capacities, we first assess semantic spaces derived from symbolic co-occurrence information in Section \ref{ssec:ppmi-sims} and then move to language model embedding-based similarity spaces in Section \ref{ssec:pythia-sims}.

\subsection{Analysis procedure}\label{ssec:semantic_sim}

NLM and human participant completions only partially overlap.
So, we implement a slight modification to RSA that allows us to compare the macro structure of language model next-word predictions and human responses independently of their overlaps for any given preamble by quantifying the changes in the structures of semantic neighborhoods.

In order to compare semantic spaces, we represent each of them as a matrix of similarities over \(W\), the set of words they have in common.
In practice, the semantic spaces we consider are instantiated as sets of vector representations, and we quantify the similarity \(s^\alpha_{i,j}\) in a given space \(\alpha\) between two words \(w_i\in W\) and \(w_j\in W\) as the cosine similarity between their vector representations \(v^\alpha_i\) and \(v^\alpha_j\) :
\begin{equation}
    s^\alpha_{i,j} = \frac{\langle v^\alpha_i \mid v^\alpha_j\rangle}{\lVert v^\alpha_i\rVert\lVert v^\alpha_j \rVert}
\end{equation}

The alignment between two semantic spaces \(\alpha\) and \(\beta\) is then computed as a function of their respective matrix representations \(S_\alpha\) and \(S_\beta\).

\subsection{Positive pointwise mutual information to assess shared response biases}\label{ssec:ppmi-sims}

The first analysis focuses on symbolic, co-occurrence based definitions of similarity. 
Consider for example cases where a model guesses both \emph{dog} and \emph{cat} for a sentence stem, and humans guess these same words for a different sentence.
According to the distributional hypothesis, so long as each data source broadly produces the same co-occurrence statistics, the model knows that it should assign high probability to words with similar meanings in the same contexts.
To probe this, we apply compute the factorization of a positive pointwise mutual information (PPMI) matrix that is derived from co-occurrence counts, which encodes distributional semantic similarities between words \citet{levy2014neural}.
We apply this procedure to the \citet{peelle2020completion} stimulus set by considering that two words \(w_i\) and \(w_j\) co-occur whenever they appear in the top-\(k\) responses\footnote{We set \(k=40\) because very few Peelle sentences elicited more than \num{40} unique completions and visual inspection of the human-NLM rank correlation (Figure \ref{fig:rank-corr-plot}) suggests that the correspondence to Peelle completions degrades rapidly. 
Consequently, much higher \(k\) would runs the risk of corrupting the semantic spaces with irrelevant words.} for a particular preamble given a particular data source.

We then compute a PPMI matrix by normalizing counts into log probabilities, subtracting the marginal log probabilities and keeping only positive values:
\begin{equation}
    \mathrm{PPMI}_{i,j} = \left[\log(p(w_i,w_j)) - \log(p(w_i)p(w_j))\right]^+
\end{equation}
We factorize this matrix using Principal Components Analysis \cite{wold1987principal} — after a row-wise normalization to produce centered and scaled values — and keep only the \(d\) dominant components, yielding \(d\)-dimensional vector representations of words.

\begin{figure}
    \centering
    \includegraphics[width=.85\linewidth]{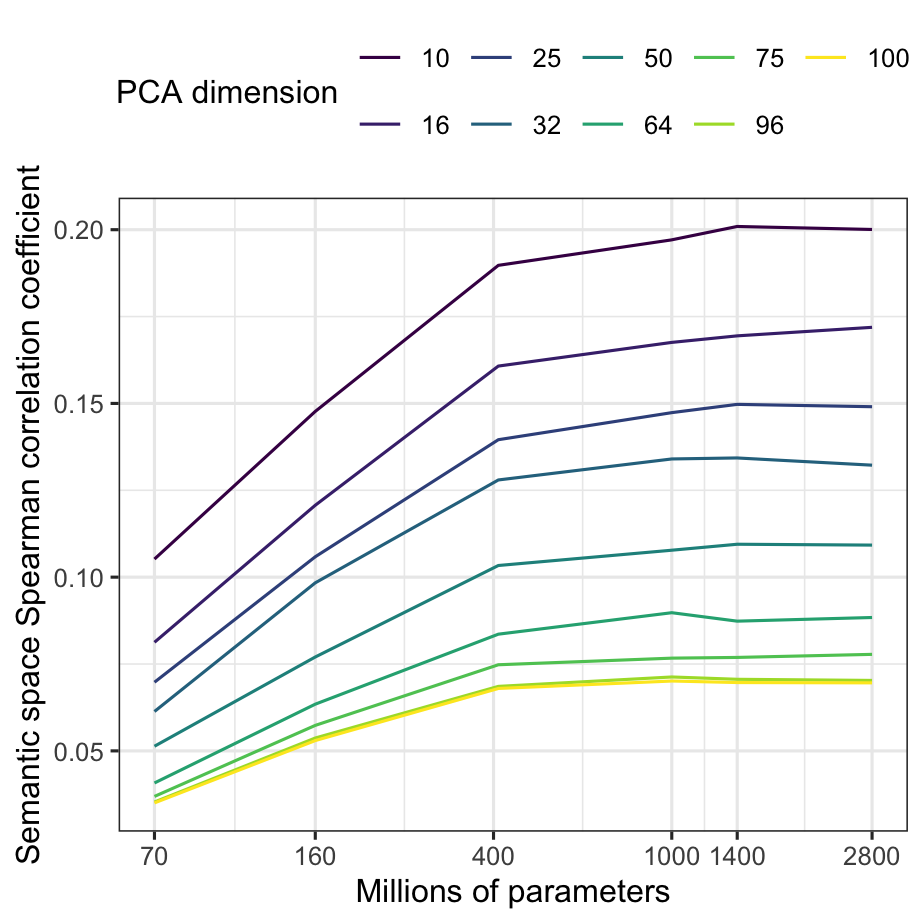}
    \caption{Correlation between model capacity, PCA dimensionality, and correlation between language model and human semantic spaces.}
    \label{fig:pca-dimensionality}
\end{figure}

We compute (cosine) similarity matrices from these vector representations (see \ref{ssec:semantic_sim}) then estimate their similarities as the Spearman correlation between their upper triangles \((s^\alpha_{i,j})_{i<j}\) and \((s^\beta_{i,j})_{i<j}\). 
We expected that this value would be high \emph{a priori} due to the degree of overlap between the human and language model responses.
All models were strongly and significantly correlated with human cloze semantic spaces, with some variation in fit based on model capacity, which generally improved as model capacity improved.
In particular, we observe that higher $d$ for PCA produced more differentiated similarity spaces, such that the size of the correlation coefficient between human and LLM completions monotonically decreased with increasing $d$ across models of all sizes  (\zcref{fig:pca-dimensionality}).

\subsection{Semantic neighborhood structure of contextual embedding representations}\label{ssec:pythia-sims}

One shortcoming of the previous analysis is that human responses can align with language model completions even in the absence of lexical overlap \cite{holtzman2021surface}.
This experiment compares the neighborhoods constructed from contextual embeddings of each NLM-generated and human-produced cloze response.
To put human and NLM-generated data sources of all sizes on equal representational footing, we use Pythia-2.8B to produce embeddings, which we extract from the last layer.
This ensures that differences in embedding quality do not detrimentally affect the comparison of semantic spaces, particularly for smaller models.
If NLMs are largely producing words that are reasonable substitutes for the same words that humans would have provided, then embeddings of those predictions should be highly similar to humans' completions. %
Additionally, if larger NLMs better capture semantic structure, then we expect semantic correspondence to improve with capacity.

Here we model semantic spaces for human and LM completions using non-contextual word representations \((v^\alpha_i)\) obtained by averaging the Pythia 2.8B contextual embeddings of all their occurrences:
\begin{equation}
    v^\alpha_i = \frac{1}{\abs{D^\alpha_i}}\sum_{c\in D^\alpha_i} h_{\mathrm{2.8B}}(w_i, c)
\end{equation}
where \(D^\alpha_i\) is the set of the preambles where word \(w_i\) appears as a completion for dataset \(\alpha\) (either human completions of completions for a given LM) and \(h_{\mathrm{2.8B}}(w_i, c)\) is the Pythia 2.8B contextual embedding of \(w_i\) given preamble \(c\).

We then define the alignment \(A_k(\alpha, \beta)\) between two spaces by their \emph{neighborhood overlap}: the degree to which each word (\enquote{pivot}) has the same neighbors in both spaces, quantified as the average Jaccard similarity between the top $k$ most similar words for a pivot in each space:
\begin{equation}
    A_k(\alpha, \beta) = \frac{1}{\abs{W}}\sum_{w\in W} \frac{\abs{V^\alpha_k(w) \cap V^\alpha_k(w)}}{\abs{V^\alpha_k(w) \cup V^\alpha_k(w)}}
\end{equation}
where \(V^\alpha_k(w)\) is the set of the \(k\) closest neighbors of word \(w\) in space \(\alpha\).

\begin{figure}
    \centering
    \includegraphics[width=.95\linewidth]{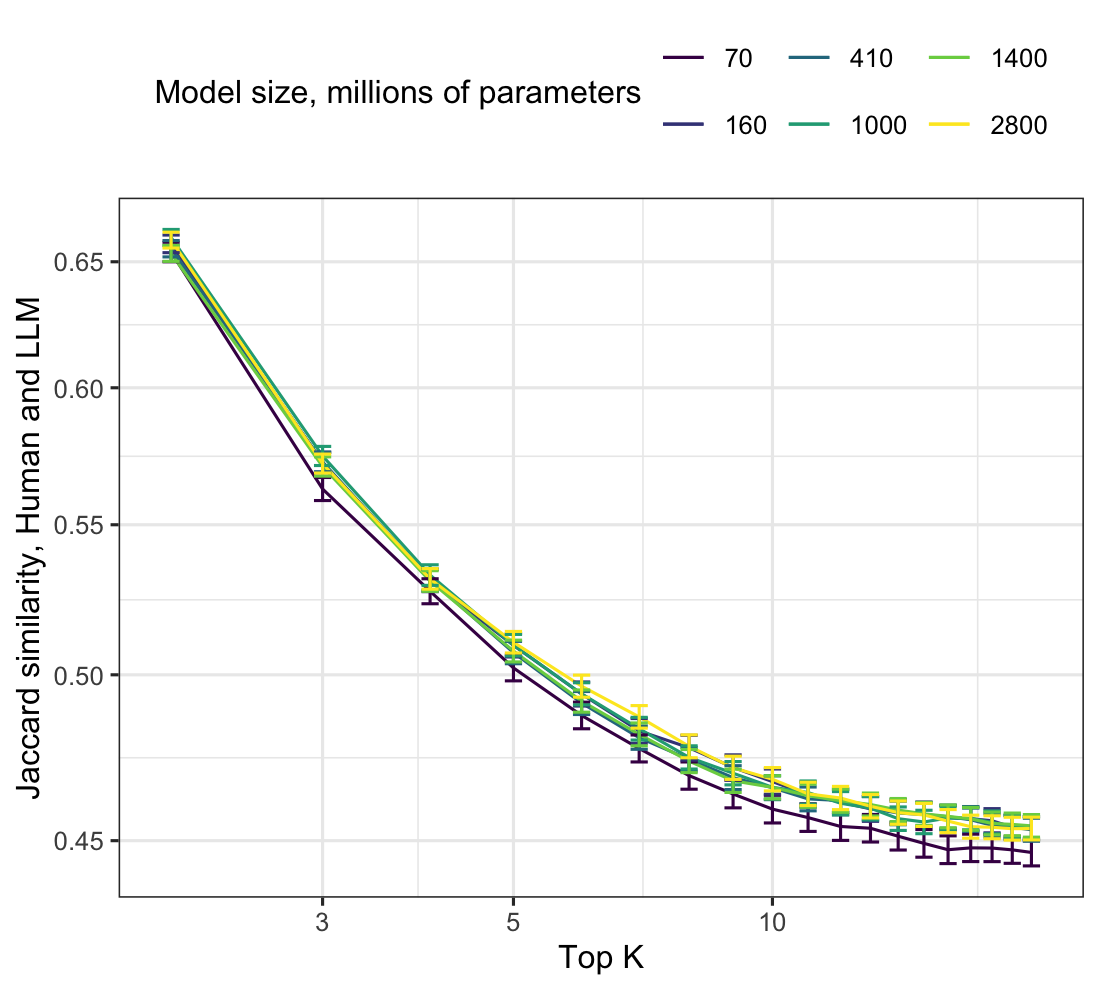}
    \caption{Neighborhood-based alignment between human and NLM completions as a function of model size and neighborhood size considered. X axis in log scale.}
    \label{fig:jaccard}
\end{figure}

For $k=20$, a generalized linear model predicting overlaps as a function of model size revealed modest improvements in overlap with greater capacity ($\beta = 0.006041$, $t(24233) = 2.540$, $p < .05$).
However, the overall alignment was very similar across models of all sizes; Pythia-70M averaged \qty{48.7}{\percent} and Pythia-2.8B averaged \qty{49.8}{\percent} neighborhood overlap.
Further analysis applying stricter cutoffs for $k$ (i.e., \numrange{2}{19}; Figure \ref{fig:jaccard}) revealed that while it is possible to produce greater semantic alignment to human data (up to \qty{66}{\percent} in the case of the top neighbor), greater model capacity did not produce significantly higher degrees of overlap.

\section{Discussion}

Probabilities extracted from NLMs correlate strongly with scores from n-gram models, reading times, and cloze probababilities.
We present strong evidence that larger models typically better explain cloze productions than smaller ones, both at the lexical and semantic levels of linguistic structure.
The consequences of this finding for computational psycholinguistics are multifaceted.
First, one possible explanation for the gap in accounting for the variety of cloze responses could be that people are not choosing words solely proportionate to their activation \cite{kumar2025lexical,meister2024towards}.
For example, some models of language production assume a type of semantic competition \cite{abdel2009semantic,harley1993phonological}.
New models could constrain or alter the next-word prediction process to explicitly account for semantic structure and better capture cloze task dynamics.
Since larger models that are trained to predict language production output, highly accurate models should be better able to capture the complexity of factors that influence word choice \cite{antonello2024predictive,oh-etal-2024-frequency,michaelovlanguage}.

Second, our findings shed light on the purported weak correspondence between human cloze productions and reading times \cite{nair2026clozinggapexploringlanguage,shain2024large,staub2025predictability,szewczyk2022context}.
Since larger models correlate better with cloze probabilities both lexically and semantically, and both are less predictive of reading times, the cloze task may engage semantics and longer-range context to a greater extent than reading tasks \cite{staub2015influence}, which may be more sensitive to narrow statistical context or perceptual features \cite{futrell2020lossy,milligan2023out,nair2026clozinggapexploringlanguage,schotter2023event}.
Language comprehension must be robust to unexpected language, and making very specific predictions may not always be optimal \cite{arora2022estimating}, so relying more on statistical regularities than semantic structure may be optimal for reading \cite{oh-etal-2024-frequency}.
Our findings complement those of \citet{nair2026clozinggapexploringlanguage}, who focus specifically on the relationship between cloze probabilities, language model probabilities, and fit to reading times.

The results of this study underscore the need to formally specify the cognitive and computational processes that are used to generate next-token predictions in both language models and humans before discarding the cloze task in favor of large language model estimates of next-token predictability.
In particular, more empirical and modeling work is needed to determine whether and how language comprehension relies on explicit, cloze-like generation abilities, and how language probabilities act as a proxy for linguistic computation \cite{staub2025predictability}.

\section{Limitations}

The present studies did not assess a very large number of language models or model architectures, which limits the generalizability of the scaling relationships we present here, which may not hold for other types. 
Additionally, our results may not generalize to systems that rely on reinforcement learning or other methods that could be used to explicitly simulate the cloze task \cite{martinez2025simulating}.
However, the work we present here suggests that work remains to be done in human-like text generation \cite{kendro2026large}.

Our analyses ultimately focus on words that are shared between data sources, and thus any biases of different models toward subword-scale predictions rather than English words may influence the semantic neighborhood structure of next-token predictions in ways that hinder alignment.
Future work should assess how these two factors interact.

Our results are likely limited cross-linguistically, as few languages have high-quality pre-trained models and large-scale cloze completion norms that can be used to study scaling relationships to cloze tasks.
Languages with more complex morphology will likely require different strategies for comparison than the ones we assess here.

\section{Ethical Considerations}
The \citet{peelle2020completion} data were gathered using crowdsourcing on Mechanical Turk. The original researchers obtained IRB approval for their research study. Some of the human responses, and many of the NLM predictions, contain profanity, sexually explicit content, or other offensive content. 
Furthermore, NLMs in general are likely to produce sexist, ableist, and racist responses even when guardrails are implemented. Researchers seeking to evaluate NLM outputs using human judgments should be aware of the potential for harmful material in these outputs.

\bibliography{tacl2021,anthology-1,anthology-2}

\end{document}